\documentclass{article}

\usepackage[nonatbib,preprint]{nips_2018}

\usepackage{amsmath}
\usepackage[utf8]{inputenc} 
\usepackage[T1]{fontenc}    
\usepackage{hyperref}       
\usepackage{url}            
\usepackage{booktabs}       
\usepackage{amsfonts}       
\usepackage{nicefrac}       
\usepackage{microtype}      

\usepackage{bbm}
\usepackage{graphicx}

\title{Trusted Neural Networks for\\ Safety-Constrained Autonomous Control}
\author{
    Shalini Ghosh\thanks{Work done while the first author was at SRI International.}\\
    Samsung Research America \\
    \texttt{shalini.ghosh@samsung.com}
    \And
    Amaury Mercier\\
    Ecole Polytechnique\\
    \texttt{amaury.mercier@polytechnique.edu}
    \And
    Dheeraj Pichapati\\
    UC San Diego \\
    \texttt{dheerajpv7@gmail.com}
    \And
    Susmit Jha\\
    SRI International\\
    \texttt{jha@csl.sri.com}
    \And
    Vinod Yegneswaran\\
    SRI International\\
    \texttt{vinod@csl.sri.com}
    \And
    Patrick Lincoln\\
    SRI International\\
    \texttt{lincoln@csl.sri.com}
    }

\newcommand{\norm}[1]{\left\lVert#1\right\rVert}

\begin{document}

\maketitle

\begin{abstract}
We propose Trusted Neural Network (TNN) models, which are deep neural
network models that satisfy safety constraints critical to the
application domain. We investigate different mechanisms for
incorporating rule-based knowledge in the form of first-order logic
constraints into a TNN model, where rules that encode safety are
accompanied by weights indicating their relative importance. This
framework allows the TNN model to learn from knowledge available in
form of data as well as logical rules.  We propose multiple approaches
for solving this problem: (a) a multi-headed model structure that
allows trade-off between satisfying logical constraints and fitting
training data in a unified training framework, and (b) creating a
constrained optimization problem and solving it in dual formulation by
posing a new constrained loss function and using a proximal gradient
descent algorithm. We demonstrate the efficacy of our TNN framework
through experiments using the open-source TORCS~\cite{BernhardCAA15}
3D simulator for self-driving cars. Experiments using our first
approach of a multi-headed TNN model, on a dataset generated by a
customized version of TORCS, show that (1) adding safety constraints
to a neural network model results in increased performance and safety,
and (2) the improvement increases with increasing importance of the
safety constraints. Experiments were also performed using the second
approach of proximal algorithm for constrained optimization --- they
demonstrate how the proposed method ensures that (1) the overall TNN
model satisfies the constraints even when the training data violates
some of the constraints, and (2) the proximal gradient descent
algorithm on the constrained objective converges faster than the
unconstrained version.
\end{abstract}

\maketitle

\section{Introduction}

Deep learning algorithms generally assume a parametric (typically
non-linear) model and fit the training data to the model to minimize
the loss between true output and model's output. Sometimes, we also
impose regularization constraints on parameters to avoid
overfitting. However, in some applications it is important to impose
additional constraints on the model output, such as constraints
enforcing safety properties inherent in the domain. For example, when
driving along a straight road, the autonomous controller for a
self-driving car should not cross a double-yellow line. Another
example could be a medical transcription tool that is faced with
ambiguity when translating text.  Domain knowledge regarding safe drug
prescription boundaries can mitigate erroneous transcriptions
indicating an unhealthy high dosage for a medication.  In this paper,
we investigate methods to add constraints on the output of a deep
neural network model as an additional objective that has to be
satisfied during training, even when the training data might violate
the constraints on some occasions. We call the resulting models
Trusted Neural Networks (TNN), as they are deep neural network models
that satisfy safety constraints critical to the application domain. In
this paper, we propose two approaches for designing TNN models:

(1) One approach is using a multi-headed model
architecture~\cite{Bagnall15}, where one ``head'' (i.e., tower of
neural network layers) fits the labeled data while another head fits
the logic constraints, and both heads have a set of common layers for
parameter sharing.  The whole model is then trained jointly,
effectively having a combined loss function with a trade-off parameter
that controls the desired importance of the logic constraints.  We run
experiments using the open-source TORCS~\cite{BernhardCAA15} 3D
simulator for self-driving cars to show that (a) adding informative
constraints to a baseline model allows it to predict safer results,
and (b) increasing the importance given to the rules gives improved
results.

(2) Second approach considers the constraints in the objective
function using a dual formulation, i.e., optimizing a new loss
function that is a weighted sum of the original loss function and a
function representing the constraints. We use proximal gradient
descent to optimize this new loss function.  We run experiments using the simulator, and demonstrate that our
constrained learning approach manages to (a) impose important
constraints on the model, and (b) guide the model in reaching the
optimal point faster.


In this paper we individually evaluate these two methods. We don't
compare them to each other, since the two methods would in the limit
generate models with similar performance (e.g., MSE) when the
constraints are known and enforced strictly. In cases where the
logical constraints are not known but needs to be learned from data,
we use the multi-headed model formulation where the logic head learns
an approximation of the constraints from data. When we know the form
of the constraints for certain, it may be better to use the
constrained learning formulation, since it can in general converge to
the optimum value faster during training.

\section{Problem Formulation}
\label{sec:problem}
Deep learning methods for training neural networks fit a function $f$ between input ($i$) and output ($o'$) and
learn the parameters (weights) $w$ so that the model's output ($o'$)
is close to true output ($o$). The learning part can be posed as an
optimization problem, where $\ell$ is a loss function:
\[
  \min_{w} \ell(o, o'), \quad \text{s.t. } o' = f(w,i).
\]
Occasionally, we add a further constraint on weights to avoid overfitting,
giving us a regularized loss function:
\[
  \min_{w} \ell(o, o') + \ell'(w), \quad \text{s.t. } o' = f(w,i)
\]
where $\ell'$ is a regularization function. To satisfy safety
constraints, we consider imposing additional constraints directly on
model's output, giving a constrained loss function:
\begin{equation}
  \label{eq:tnn}
  \min_{w} \ell(o,o'), \quad \text{s.t. } o' = f(w,i) \  \mbox{and} \ g(o') \le 0
\end{equation}
where $g$ is a function, possibly in first-order logic,
specifying the safety constraints. In this paper, we study two
possible ways of training trusted neural network (TNN) models by
solving the optimization problem given in Equation~\ref{eq:tnn}.

\section{Trusted Neural Networks}
\label{sec:methods}

The first approach for training TNN models uses a multi-headed model
architecture to learn from two different sources of knowledge: 
data-based and rule-based.

\subsection{Multi-headed models}
\begin{figure}[htbp]
    \centering
    \begin{minipage}{0.5\textwidth}
        \centering
        \includegraphics[width=0.5\linewidth]{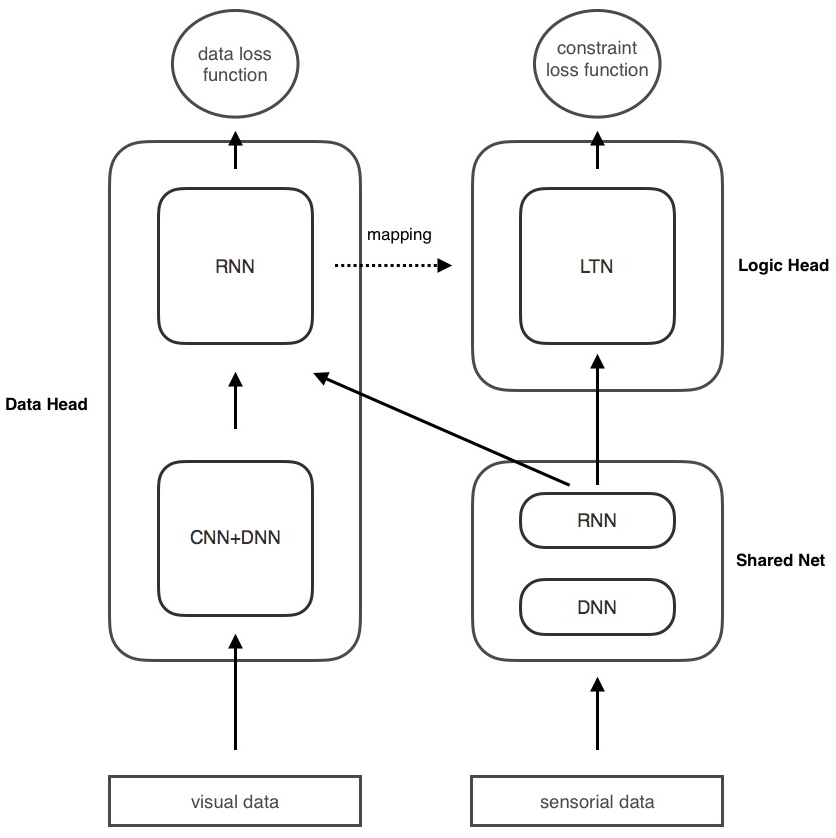}
        \caption{The multi-headed TNN model.}
        \label{fig:multiheaded_model_car}
    \end{minipage}%
    \begin{minipage}{0.5\textwidth}
        \centering
        \includegraphics[width=0.6\linewidth]{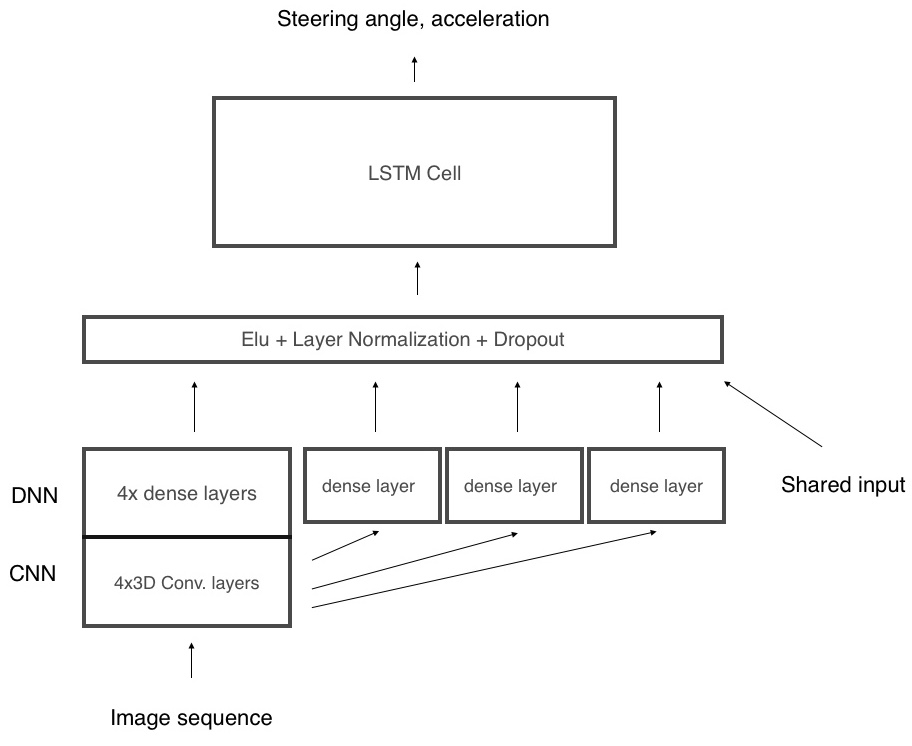}
        \caption{The data head.}
        \label{fig:data_head}
    \end{minipage}
\end{figure}

In the multi-headed architecture, each of the model's heads has its
own objective function and is adapted to the problem they have to
solve --- they share a neural network (set of layers), called
the shared net. One example of a multi-headed network is given
in Figure~\ref{fig:multiheaded_model_car}; let the parameters of the
head 1, the head 2 and the shared network be $p_{1}$, $p_{2}$
and $p_{s}$ respectively. Here, head 1 has the objective function
$O_{1}(p_{1}, p_{s})$ and head 2 has the objective function
$O_{2}(p_{2}, p_{s})$. The model is then jointly trained with the
combined loss function, $O_{1}(p_{1}, p_{s}) + \lambda O_{2}(p_{2},
p_{s})$ where $\lambda \in \mathbb{R^+}$ is a parameter fixing a
trade-off --- it determines the importance given to the
rule-based knowledge source.

Note that if we need the constraints to be strictly enforced, we can
set a very high value of $\lambda$. Having soft constraint
satisfaction using the Lagrangian formulation allows us to enforce the
constraints with different levels of strictness, which can be enforced
based on the domain requirements.

\subsubsection{Network on logical constraints}

To fit logic rules in the logic head, we chose to use and extend Logic
Tensor Networks~\cite{serafini2016logic}, i.e., LTN. In LTNs, we
define a language using first-order logic $\mathcal{L}(\mathcal{C},
\mathcal{F}, \mathcal{P})$, where $\mathcal{C}$ is the set of
constants (data points), $\mathcal{F}$ the function symbols, and
$\mathcal{P}$ the predicate symbols. The goal of LTNs is to typically
learn the function $g$ in Equation~\ref{eq:tnn}, from both data and
rule-based knowledge (written as first-order logic rules).  To do
this, a grounding is defined on the language $\mathcal{L}$, mapping
logical propositions to their truth values in $[0,1]$. Constants are
mapped to numerical vectors; the symbols we want to learn are mapped
to functions whose parameters are learned using gradient descent, to
maximize an objective function that is the conjunction of clauses
defined over the data points. In the TNN model, we jointly train a
network (the data head) on the data and a LTN (the logic head) on the
rules, using a combined loss function.

The functional operators in the LTN are mapped to small deep
regression neural networks.  The LTN has to effectively encode logical
functions in a neural network, for which it uses real
logic~\cite{serafini2016logic}.  Real logic is defined as follows.
Let $\mathcal{G}$ be the grounding used in the real logic framework,
and $f$ a real unary functional operator $f : \mathcal{C} \rightarrow
\mathbb{R}^{p}$ that will be mapped to the function
$\mathcal{G}(f):\mathbb{R}^{d} \rightarrow \mathbb{R}^{p}.$ The
definition of the grounding $\mathcal{G}$ is extended to a new class
of literals using the functional operator $f$; mainly, for a data
point $x \in \mathcal{C}$ and $c \in f(\mathcal{C}),$ $f(x)=c$,
$f(x)\ge c$ and $f(x)\le c,$ are:
\[ 
\mathcal{G}(f(x)=c) \in [0,1], \quad 
\mathcal{G}(f(x)=c) = 1 \iff f(x)=c,
\] \[
\mathcal{G}(f(x) \leq c) \in [0,1], \quad 
\mathcal{G}(f(x) \leq c) = 1 \iff f(x) \leq c,
\] \[
\mathcal{G}(f(x) \geq c) \in [0,1], \quad 
\mathcal{G}(f(x) \geq c) = 1 \iff f(x) \geq c.
\]
Possibilities include using functions such as :
\begin{equation}
\label{gaussian}
\mathcal{G}(f(x)=c) = 1 - exp(||f(x) - c||^{2})
\end{equation}
and if $f(\mathcal{C})$ is bounded with a diameter $\delta(\mathcal{C})$
\begin{equation}
\label{squared_difference}
\mathcal{G}(f(x)=c) = 1 - \left|\left|\frac{f(x) - c}{\delta(\mathcal{C})}\right|\right|^{2}, \  \mbox{or}
\end{equation}
\begin{equation}
\label{absolute_difference}
\mathcal{G}(f(x)=c) = 1 - \left|\frac{f(x) - c}{\delta(\mathcal{C})}\right|.
\end{equation}

The function in Equation~\eqref{gaussian} provides very low gradients
outside of the truth region, while
Equation~\eqref{absolute_difference} is non-differentiable on 0; we
decided to use Equation~\eqref{squared_difference}, standard in the
field of regression, as the functional operators considered have
defined bounds.
Adapting to the other literals :

\[ \mathcal{G}(f(x) \leq c) = 1 - \left|\left|\max\left(\frac{f(x) - c}{\delta(\mathcal{C})}, 0\right)\right|\right|^{2}, \quad
 \mathcal{G}(f(x) \geq c) = 1 - \left|\left|\min\left(\frac{f(x) - c}{\delta(\mathcal{C})}, 0\right)\right|\right|^{2}. \]


To implement bounded regression neural networks, we use a $tanh$
activation function on the output layer and rescale the output with
the desired mean and range vectors.  Assume $f(\mathcal{C}) \subset
[a_{0} ; b_{0}] \times [a_{1} ; b_{1}] \times ... \times [a_{p} ;
  b_{p}] $ with $a_{0},...,a_{p},b_{0},...,b_{p} \in \mathbb{R}.$ Define
$m = \frac{1}{2}(a_{0} + b_{0},a_{1} + b_{1}, ..., a_{p} + b_{p})$ and
$r = \frac{1}{2}(b_{0} - a_{0}, b_{1} - a_{1}, ..., b_{p} -
a_{p})$. Then the output of the neural network is $m + r \odot
tanh(y)$, where $y$ is the output of the last layer of the neural
network under consideration. \\

The model learns from the rules provided that any precondition of the
rule is present in the dataset. Otherwise, the rule is considered as
satisfied by the model.

\subsection{Constrained Neural Network Learning}

In our second approach, we solve the problem in Equation~\ref{eq:tnn}
by projecting the optimization problem in dual space as:
\begin{equation}
  \label{eq:dual}
  \min_w \ell(o,o')+\lambda g(o') \quad
  \text{s.t. } o' = f(w,i), \quad \quad \lambda \ge 0
\end{equation}

Using grid search we can explore over possible values of  hyperparameter $\lambda$ 
and find the best value. Greater values of $\lambda$ give more
significance to the constraint function. We solve the above
optimization problem~(\ref{eq:dual}) using stochastic proximal
gradient descent~\cite{Nesterov07,DuchiY09}.
Proximal gradient descent updates the parameters in two steps:
\begin{align*}
  w_{t+1/2} &= w_t -\eta_t\partial \ell(w_t)\\
  w_{t+1}   &= \arg\min_{w}\left(\frac12\norm{w-w_{t+1/2}}^2+\eta_t \lambda g(w)\right).
\end{align*}
where in step 1 parameters move along in the direction of gradient of 
$\ell$ and in step 2 we do proximal mapping of $g$.

Notice that first step is regular gradient descent and can be
implemented using backpropagation. The second step is the proximal
mapping and for simple $g$ functions, we can have closed form proximal
mappings. In other cases, we can use the backpropagation algorithm to
find an approximation.

\section{Related Work}
\label{sec:relatedwork}

Machine learning (ML) has a rich history of learning under
constraints~\cite{dietterich:thesis,miller:94} --- different types of
learning algorithms have been proposed for handling various kinds of
constraints. Propositional constraints on size~\cite{bar-hillel:2005},
monotonicity~\cite{kotlowski:2009}, time and
ordering~\cite{laxton:2007}, etc. have been incorporated into learning
algorithms using constrained optimization~\cite{bertsekas:book} or
constraint programming~\cite{raedt:aaai10}, while first order logic
constraints have also been introduced into ML
models~\cite{mei:icml04,richardson:2006}.

Hu et al.~\cite{Hu2016} incorporated logic rules into a DNN or RNN
using a teacher/student network --- they iteratively train a student
network on the data and project the student network onto the set of
given rules to get the teacher network. Towell et
al.~\cite{Towell1994} have also encoded propositional rules into
neural networks. Our approaches, based on multi-headed NNs or
constrained loss functions, are more flexible, allowing a fine-tuned
trade-off between rule constraints and data knowledge, and enables
constrained-guided data-efficient learning in a novel way. Ghosh et
al.~\cite{ghosh:aaai2017} proposed a methodology to ensure that
probabilistic models learned from data can also satisfy safety
constraints expressed as first-order logic rules.

Proximal gradient descent has been used to optimize convex functions
in ML frameworks other than neural
networks~\cite{BoydNEBJ11,flaam1996equilibrium,combettes2011proximal,liu:nips12,thomas:pnac13}. Proximal
gradient descent has been used in neural network frameworks for
regularization~\cite{DuchiY09}. Constrained optimization in neural
networks has also been studied~\cite{serafini2016logic}, where Real
Logic (first order logic constraints which have truth value in range
$[0,1]$ and real number semantics) have been integrated with the
neural networks framework.

\section{Experimental Evaluation}
\label{sec:exp}

\subsection{Multi-headed TNN model}

We demonstrate the usefulness of the multi-headed TNN model using
experiments with an autonomous car controller. We implemented our
neural network models in Tensorflow~\cite{AbadiAPEZCGAJM16}.

\subsubsection{Car controller}

Car control is an example ML problem where logic constraints (e.g.,
regarding safety) could be used in addition to training data from
sensors to produce better and safer models.  We developed a customized
version of the open-source TORCS~\cite{BernhardCAA15} 3D car simulator
for self-driving cars, which is used for car racing competitions. It
gives access to a variety of numeric sensors to help control the car
--- distance to the edge of the track or to the closest opponent in a
set of directions, speed, distance and angle to the center of the
track. The problem is to control a set of 5 variables: acceleration,
brake, clutch, gear, and steering angle. In our customized version of
TORCS, the ML model has also access to the front camera of the car. We
assume there are no other cars on the road.  The driving state we
consider is a vector $(I, S, A)$ where $I$ is the front camera image,
$S$ is the numerical vector of the sensors, and $A$ is the action
vector.


\subsubsection{Dataset}
For generating the datasets, we considered a simple car controller with
a fixed objective speed and position on the track.
The behavior of the vehicle in the simulator is recorded on 10
different tracks
included in TORCS, at a framerate of 17 frames per second. The
resulting dataset includes image from the front camera, sensorial time
trace, and output data.
We create two datasets for experimentation.
Dataset A is generated based on a deterministic driving pattern, while
in Dataset B we introduce some randomness in parameters like track
objective. Dataset B has a more diverse driving behavior than dataset
A\footnote{We will be releasing the code and these
  datasets to the research community soon. 
  }


\subsubsection{Multi-headed model}
 
The full multi-headed TNN model, depicted
Fig.~\ref{fig:multiheaded_model_car}, is jointly trained with a
combined loss function:
 \[ totalLoss = dataLoss + \lambda_{loss}.logicLoss \]
where $dataLoss$ is the loss function relative to the data head, and
$logicLoss$ is the one relative to the logic head.

\subsubsection{Shared net}
The shared net
is a Long short-term memory (LSTM) cell~\cite{hochreiter:1997} built
on top of a small deep neural network with two ReLu-activated
layers~\cite{nair:icml10}.  The input of the shared net is the time
trace of the sensorial data, and its output is connected to both data
and logic heads.

\subsubsection{Data head}
The data head is the baseline for our problem. In the Udacity
challenge \#2~\cite{udacity2}, where competitors had to predict
steering angle from a sequence of image, the winner used a sequential
model (RNN) based on a customized LSTM cell on top of a deep
convolutional neural net (CNN)~\cite{lecun:1995} --- we considered
that as our baseline model.

The goal of the data head is: given a sequence of images and sensorial
vectors, predict the steering angle sequence. We add the acceleration
to the predicted variables to force the model to learn more precisely
the momentum of the car. The data head, as depicted in
Fig.~\ref{fig:data_head}, is an LSTM cell whose input is the
concatenation of the output of two nets --- one deep convolutional
neural net made of 4 ReLu-activated convolutional layers and 4
ReLu-activated dense layers, and the output of the shared
net.
The data head uses the mean square error (MSE) loss function.


\subsubsection{Logic head}
\begin{figure}[hbtp]
    \centering
    \begin{minipage}{.48\textwidth}
        \centering
        \includegraphics[width=0.5\linewidth]{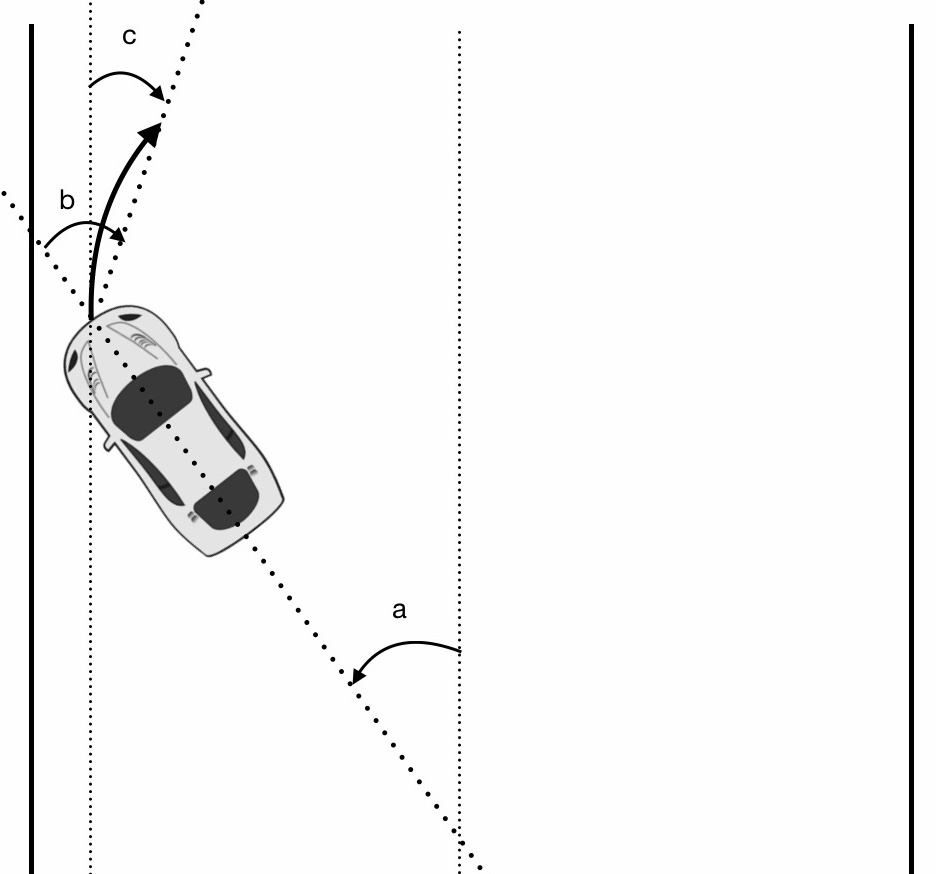}
        \caption{Geometric variables in the TORCS setup.}
        \label{fig:car_rule}
    \end{minipage}
    \begin{minipage}{.48\textwidth}
        \centering
        \includegraphics[width=0.5\linewidth]{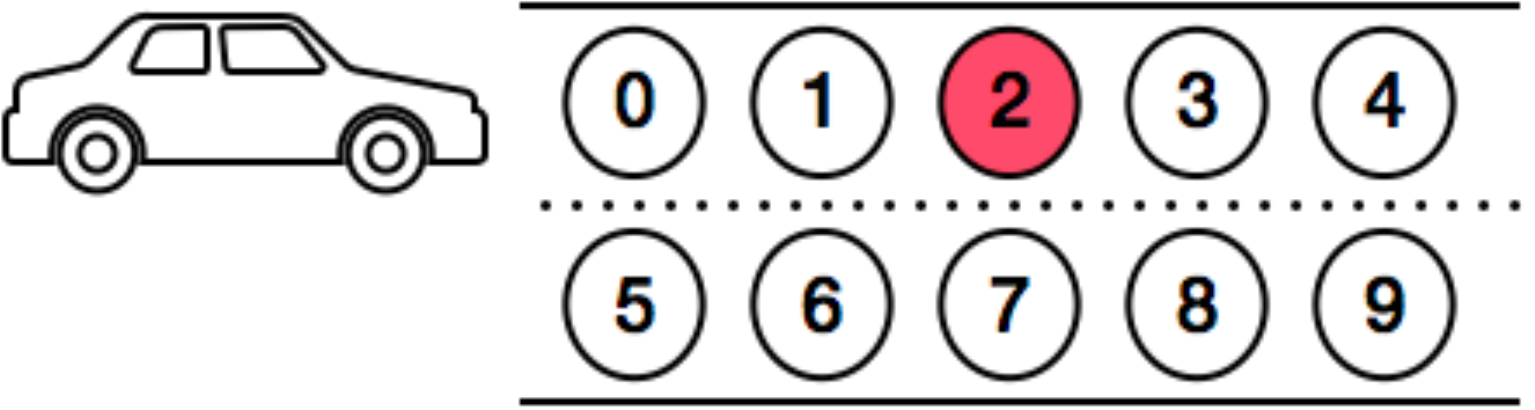}
        \caption{State map with obstacle at state 2.}
        \label{fig:road}
    \end{minipage}
\end{figure}

The logic head uses the output of the data head to map the data
points: a data point is  mapped to the vector $(S, dataHeadOutput)$,
where S is the sensorial time trace.
Fig.~\ref{fig:car_rule} explains how we design a rule for our
experiment. Let $a, b, c$ be the oriented angles depicted in
Fig.~\ref{fig:car_rule}. We have:
\[
\begin{cases}
a = - S['angleToTrack'], \\
b = S['steeringAngle'], \\
c=a+b=S['steeringAngle']-S['angleToTrack']
\end{cases}
\]

Intuitively, if the car is close to the right edge of the road, we
want $c>0$ ; if the car is on the left edge of the road, we want
$c<0$. This gives the following set of two rules :
\[
\begin{cases}
if S['trackPosition'] < -0.75,  \quad S['steeringAngle']  - S['angleToTrack'] > 0 \\
if S['trackPosition'] > +0.75,  \quad S['steeringAngle']  - S['angleToTrack'] < 0
\end{cases}
\]

To implement these rules, we define a functional operator
$steeringAngle: \mathcal{C} \rightarrow [-1,1]$ that is mapped to a
small bounded deep neural network. The rules are encoded using the
real logic framework and define, for a given batch of data points and
for each rule, the truth values $ruleClauseLeft$ and
$ruleClauseRight$. We add the clauses stating that the output of
$steeringAngle$ for each element of the batch has to be the ground
truth value, which defines a truth value $steeringClause$. These
clauses are aggregated using a weighted sum --- the weight for
$steeringClause$ is fixed to 1, and the weight for each rule is a
parameter to define using cross validation: $w_{rule}.$ As a result,
we have:
\begin{align*}
 logicLoss = 1 - \frac{1}{1 + 2w_{rule}}(steeringClause 
  + w_{rule}(ruleClauseLeft + ruleClauseRight)).
\end{align*}

\subsubsection{Evaluation metrics}

We evaluate the model against the baseline (data head without the
logic head) on two different metrics:

(1) The first metric is the mean square error against the ground truth
for steering angle prediction. It evaluates how well knowledge from 
data is integrated into the model.

(2) The second metric uses the logic head rules. The goal is to
evaluate how well the rules have been learned by the model. For a
sequence of driving states $((I_{1}, S_{1}, A_{1}),...,(I_{L}, S_{L},
A_{L})),$ where $A_{1}, .., A_{L}$ is the prediction of the model, let
$c_{i} = S_{i}['steeringAngle']-S_{i}['angleToTrack']$ for each
state. Note that $c_{i}$ should be negative when
$S_{i}['trackPosition'] \geq 0.75$ (car is near the left edge of the
road) and positive when $S_{i}['trackPosition'] \leq -0.75$ (car is
near the right edge of the road). We define the danger metric of a TML
model as:
\[
  danger =  \sum_{i=1}^{L} violationCost_i
\]
\noindent where $violationCost_i$ is the cost of the $i^{th}$
datapoint violating a safety constraint. For our current dataset and
constraints, we consider the cost to be proportional to $|c_i|$ if the
$i^{th}$ datapoint violates a constraint, giving us this danger
measure:
\begin{align*}
  danger =  \sum_{\substack{i=1 \\ S_{i}['trackPosition'] \geq 0.75}}^{L} max(c_{i}, 0)   + \sum_{\substack{i=1 \\ S_{i}['trackPosition'] \leq -0.75}}^{L} -min(c_{i}, 0)
\end{align*}

\subsubsection{Parameter selection}

Considering $steeringLoss = 1 - steeringClause$ and $ruleLoss = 2 -
ruleClauseLeft - ruleClauseRight$:
\begin{align*}
totalLoss = dataLoss + \frac{\lambda_{loss}}{w_{rule}}.steeringLoss  + \lambda_{loss}.ruleLoss
\end{align*}

To understand the influence of the $\lambda_{loss}$ parameter, we
fix $w_{rule}$ and vary $\lambda_{loss}$ on both datasets, and check
how the MSE and danger metrics vary for the data head and logic head
networks. The reported results are computed using 5-fold
cross-validation.


\subsubsection{Results}

\begin{figure}[hbtp]
    \centering
    \begin{minipage}{.48\textwidth}
        \centering
        \includegraphics[width=0.7\linewidth]{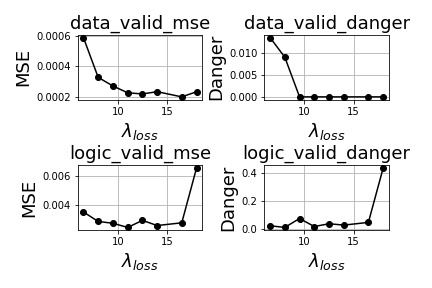}
        \caption{Metrics on Dataset A.}
        \label{fig:results_0}
    \end{minipage}
    \begin{minipage}{.48\textwidth}
        \centering
        \includegraphics[width=0.7\linewidth]{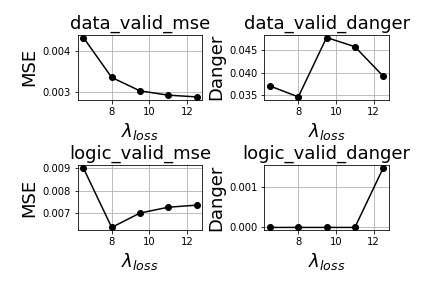}
        \caption{Metrics on Dataset B.}
        \label{fig:results_1}
    \end{minipage}
\end{figure}

Fig.~\ref{fig:results_0} and \ref{fig:results_1} show the influence of
the $\lambda_{loss}$ parameter on both metrics relative to the output
of each head (data head and logic head), for Datasets A and B
respectively. On Dataset A (Fig.~\ref{fig:results_0}), which is easier
to fit, increasing $\lambda_{loss}$ improves the data head in terms of
lowering both the MSE and danger metrics. However, when the
$\lambda_{loss}$ is too high, training becomes worse (especially for
the logic head), since the rule constraints become too strict to
enforce. On Dataset B (Fig.~\ref{fig:results_1}), which is harder to
fit, the data head MSE is initially higher but the logic head danger
is lower --- this means that even if the data is harder to fit, its
completeness (i.e., more varied distribution of trackPosition) allows
the overall model to better learn the rule. As $\lambda_{loss}$ is
increased, the MSE of the data head improves but when it becomes too
high, the logic head loses in danger metric since the rule constraints
become too strict to enforce.

Overall, we see that adding the rules to the neural network model
results in increased performance (lower MSE metric) and safety (lower
danger metric), and the improvement increases with increasing
importance of the safety constraints, i.e., value of $\lambda_{loss}$
(unless its value becomes too high).

\subsection{Constrained Optimization for TNN}

We demonstrate the usefulness of our constrained optimization approach
to TNN using two experiments that involve the autonomous car
controller.

\subsubsection{Obstacle experiment}

In our first experiment, we consider a car moving on a left lane on a
two road lane with an obstacle ahead in the lane. The ideal action for
car would be to move to right lane, pass the obstacle on right lane
and return back to left lane. We can express the situation as a state,
action pair where state refers to the location of the car and action
refers to steering left, steering right or moving straight. In
Fig.~\ref{fig:road}, states 0 to 4 are in left lane (in straight
line) and states 5 to 9 are in right lane. The obstacle is present at
state 2 so car should steer to right lane before state 2 and steer
back to left lane after state 7. The possible actions are 0 $:$ go
straight, 1 $:$ go right and 2 $:$ go left.  The optimal state-action
pairs are as follows: 0 $\rightarrow$ 1, 1 $\rightarrow$ 1, 2
$\rightarrow$ 0, 3 $\rightarrow$ 0, 4 $\rightarrow$ 0, 5 $\rightarrow$
0, 6 $\rightarrow$ 0 , 7 $\rightarrow$ 0, 8 $\rightarrow$ 2, 9
$\rightarrow$ 2.

To generate noisy dataset, we
consider a probabilistic version of  Markov Decision Process (MDP) version of this state-action
map: at each state the car takes the optimal action with probability
$0.8$ and each other action with probability $0.1$ each. Whenever the
car runs into left curb (left turn from left lane) or runs into right
curb (right turn from right lane) or runs into state 2, we restart the
trace from state 0.

We considered a single hidden layer neural network with 10 dimensional
input $x$ as the input layer. State $i$ is represented by
$\mathbbm{1}_i$, a vector with $1$ in $i$th position and 0s every
other position. First hidden layer consists 5 neurons with sigmoid
being the activation function. Next layer consists of 3 neurons with
output $y$ being the softmax of these 3 neurons. Output represents the
probability of each action i.e., $y(i)$ represents the probability of
action $i$. The loss function considered is 
the cross entropy loss. We
consider the constraint that car should not take left turn from left
lane and right turn from right lane. In other words, if $x = 1_i$ with
$i \le 4$ then $p(y(2)) = 0$ and if $x = 1_i$ with $i > 4$ then
$p(y(1)) = 0$.

We run two sets of experiments: 1) without constraints, and 2) with
constraints. As expected, when we train neural networks using
unconstrained method, the model does not supress the probability of
bad turns. For example, for state $0$, the model still outputs a
probability of around 0.1 for left turn, which is dangerous.  In the
second set of experiments, we use constrained training. We can get an
optimal value of the hyperparamter $\lambda$ by doing gridsearch, but
in this experiment we set it to a constant value of 1.0. Our method
suppresses the probability of bad turns to a low value. For example,
for state $0$, the model outputs a probability of $0.0093269$. Note
that this value can be reduced further by increasing $\lambda$.

\begin{figure}[hbtp]
    \centering
    \begin{minipage}{.48\textwidth}
        \centering
        \includegraphics[width=2in]{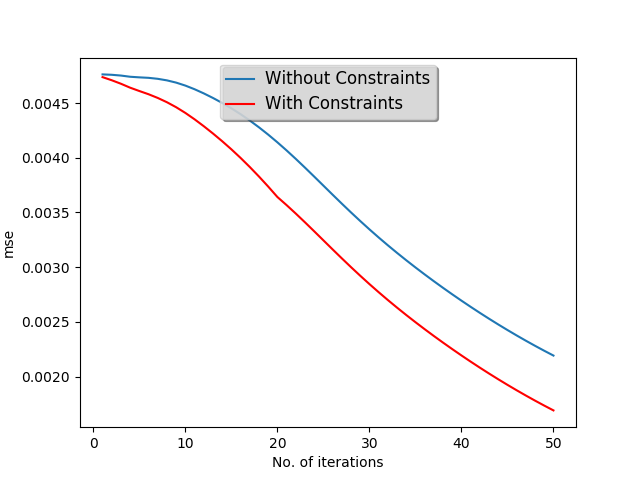}
        \caption{Convergence rate on Dataset A}
        \label{fig:A}
    \end{minipage}
    \hfill
    \begin{minipage}{.48\textwidth}
        \centering
        \includegraphics[width=2in]{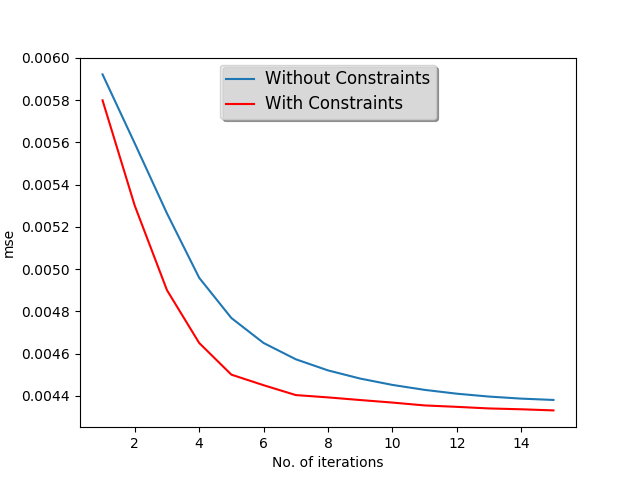}
        \caption{Convergence rate on Dataset B}
        \label{fig:B}
    \end{minipage}
\end{figure}

\subsubsection{TORCS experiment}

We also ran experiments on the TORCS datasets A and B.  We fit a
neural network to learn the function from sensorial data to steering
angle, using a single hidden layer neural network with sensorial data
from past 2 seconds as input, 10 hidden nodes and sigmoid as the
activation function and output being the steer angle. We consider mean
square error (MSE) as the loss function.  We consider the same safety
constraints that were used in the logic head of the multi-headed TNN
model in the previous experiment. Here, the training data is not too
noisy to result in dangerous steering angle but we still show that
constraints can help in reaching optimal faster.

As shown in Figs.~\ref{fig:A} and~\ref{fig:B}, the MSE of the
constrained method is less than that of the unconstrained one, and it
also converges faster. Note that because of the more noisy nature of
Dataset B, the MSE for it does not converge to 0.

\section{Conclusion and Future Work}
\label{sec:conclusion}

We introduced Trusted Neural Networks (TNNs), where we investigated
two methods of incorporating logical constraints into neural
networks. We demonstrated that TNNs enable higher trustworthiness in
deep learning systems, and can improve efficiency of training.
Specifically, we demonstrated these benefits using data generated with
the open-source TORCS 3D car simulator. The novelty of TNN is not in
designing a new model -- the novelty is in using the multi-headed
architecture for getting safety-aware ML predictions, and in using the
proximal updates for getting faster convergence in this constrained setting.

Future directions of research in this vein can include application of
TNNs to other high-consequence domains such as healthcare and
industrial control.  We would also like to theoretically study the
practical limits of trustworthiness achievable with TNN models, e.g.,
focus on deriving bounds on the generalization error for constrained
functions that can be represented by the TNN models.

\section*{Acknowledgment}
The authors would like to thank Dr. Ashish Tiwari for his valuable support and feedback regarding this work. This project was partially funded by the National Science Foundation (NSF) under Award Number CNS-1314956. Any opinions, findings, and conclusions or recommendations expressed in this material are those of the author(s) and do not necessarily reflect the views of NSF.

\bibliographystyle{abbrv}
\bibliography{refs} 
\end{document}